\documentclass[acmtog,nonacm]{acmart}

\copyrightyear{2024}
\acmYear{2024}
\setcopyright{rightsretained}
\acmConference[SA Conference Papers '24]{SIGGRAPH Asia 2024 Conference Papers}{December 3--6, 2024}{Tokyo, Japan}
\acmBooktitle{SIGGRAPH Asia 2024 Conference Papers (SA Conference Papers '24), December 3--6, 2024, Tokyo, Japan}\acmDOI{10.1145/3680528.3687664}
\acmISBN{979-8-4007-1131-2/24/12}

\acmSubmissionID{860}

\usepackage{booktabs} 
\usepackage{enumitem}
\usepackage[ruled]{algorithm2e} 
\usepackage[size=small]{caption}

\SetAlFnt{\small}
\SetAlCapFnt{\small}
\SetAlCapNameFnt{\small}
\SetAlCapHSkip{0pt}

\makeatletter
\renewcommand{\@authorsaddresses}{}
\makeatother

\newcommand{\red}[1]{{\color{red}#1}}
\newcommand{\todo}[1]{{\color{red}#1}}
\newcommand{\TODO}[1]{\textbf{\color{red}[TODO: #1]}}
\newcommand{\sz}[1]{\textcolor{violet}{[shiranz: #1]}}
\newcommand{\roni}[1]{\textcolor{green}{[roni: #1]}}
\newcommand{\brian}[1]{\textcolor{blue}{[Brian: #1]}}
\newcommand{\tali}[1]{\textcolor{magenta}{[Tali: #1]}}
\newcommand{\erika}[1]{\textcolor{brown}{[Erika: #1]}}
\newcommand{\jingwei}[1]{\textcolor{orange}{[Jingwei: #1]}}
\newcommand{\forrester}[1]{\textcolor{purple}{[Forrester: #1]}}
\newcommand{\aleks}[1]{\textcolor{red}{[Aleks: #1]}}

\renewcommand{\red}[1]{}
\renewcommand{\todo}[1]{}
\renewcommand{\TODO}[1]{}
\renewcommand{\sz}[1]{}
\renewcommand{\roni}[1]{}
\renewcommand{\brian}[1]{}
\renewcommand{\tali}[1]{}
\renewcommand{\erika}[1]{}
\renewcommand{\jingwei}[1]{}
\renewcommand{\forrester}[1]{}
\renewcommand{\aleks}[1]{}

\newcommand{\ignore}[1]{}

\definecolor{R1}{RGB}{239,176,161}

\definecolor{R2}{RGB}{75,163,195}

\definecolor{R3}{RGB}{214,40,40} 

\definecolor{R4}{RGB}{61,163,93}

\usepackage{xspace}
\usepackage{colortbl}

\makeatletter
\DeclareRobustCommand\onedot{\futurelet\@let@token\@onedot}
\def\@onedot{\ifx\@let@token.\else.\null\fi\xspace}

\def\eg{\emph{e.g}\onedot} 
\def\ie{\emph{i.e}\onedot} 

\definecolor{URL}{RGB}{239,176,161}
\newcommand{\URLColor}[1]{\textcolor{URL}{#1}}

\begin{document}
\title{VidPanos: Generative Panoramic Videos from Casual Panning Videos}

\author{Jingwei Ma}
\email{jingweim@cs.washington.edu}
\affiliation{%
 \institution{University of Washington}
 \country{USA}}
\affiliation{%
\institution{Google DeepMind}
\country{USA}}

\author{Erika Lu}
\affiliation{%
  \institution{Google DeepMind}
  \city{New York}
  \country{USA}}
\email{erikalu@google.com}

\author{Roni Paiss}
\affiliation{%
  \institution{Google DeepMind}
  \city{Tel Aviv}
  \country{Israel}
}
\email{ronipaiss@google.com}

\author{Shiran Zada}
\affiliation{%
  \institution{Google DeepMind}
  \city{Tel Aviv}
  \country{Israel}
}
\email{shiranz@google.com}

\author{Aleksander Holynski}
\affiliation{%
  \institution{UC Berkeley}
  \country{USA}
}
\affiliation{%
  \institution{Google DeepMind}
  \city{San Francisco}
  \country{USA}
}
\email{holynski@google.com}

\author{Tali Dekel}
\affiliation{%
  \institution{Weitzmann Institute of Science}
  \city{Rehovot}
  \country{Israel}
}
\affiliation{%
\institution{Google DeepMind}
\country{Israel}}
\email{tdekel@google.com}

\author{Brian Curless}
\affiliation{%
  \institution{University of Washington}
  \city{Seattle}
  \country{USA}
}
\affiliation{%
\institution{Google DeepMind}
\country{USA}}
\email{curless@cs.washington.edu}

\author{Michael Rubinstein}
\affiliation{%
  \institution{Google DeepMind}
  \city{Cambridge}
  \country{USA}
}
\email{mrub@google.com}

\author{Forrester Cole}
\affiliation{%
  \institution{Google DeepMind}
  \city{Cambridge}
  \country{USA}
}
\email{fcole@google.com}

\begin{abstract}
Panoramic image stitching provides a unified, wide-angle view of a scene that extends beyond the camera's field of view. Stitching frames of a panning video into a panoramic photograph is a well-understood problem for stationary scenes, but when objects are moving, a still panorama cannot capture the scene. 
We present a method for synthesizing a panoramic video from a casually-captured panning video, as if the original video were captured with a wide-angle camera.  
We pose panorama synthesis as a space-time outpainting problem, where we aim to create a full panoramic video of the same length as the input video. Consistent completion of the space-time volume requires a powerful, realistic prior over video content and motion, for which we adapt generative video models. Existing generative models do not, however, immediately extend to panorama completion, as we show. We instead apply video generation as a component of our panorama synthesis system, and demonstrate how to exploit the strengths of the models while minimizing their limitations. Our system can create video panoramas for a range of in-the-wild scenes including people, vehicles, and flowing water, as well as stationary background features. Project page at: \URLColor{\url{https://vidpanos.github.io}}.
\sz{Not related to the abstract. According to Miki, the copyright footnote can be removed for the SIGGRAPH submission. It will need to be added later for the camera ready version}
\end{abstract}


\begin{teaserfigure}
    \includegraphics[width=\textwidth]
    {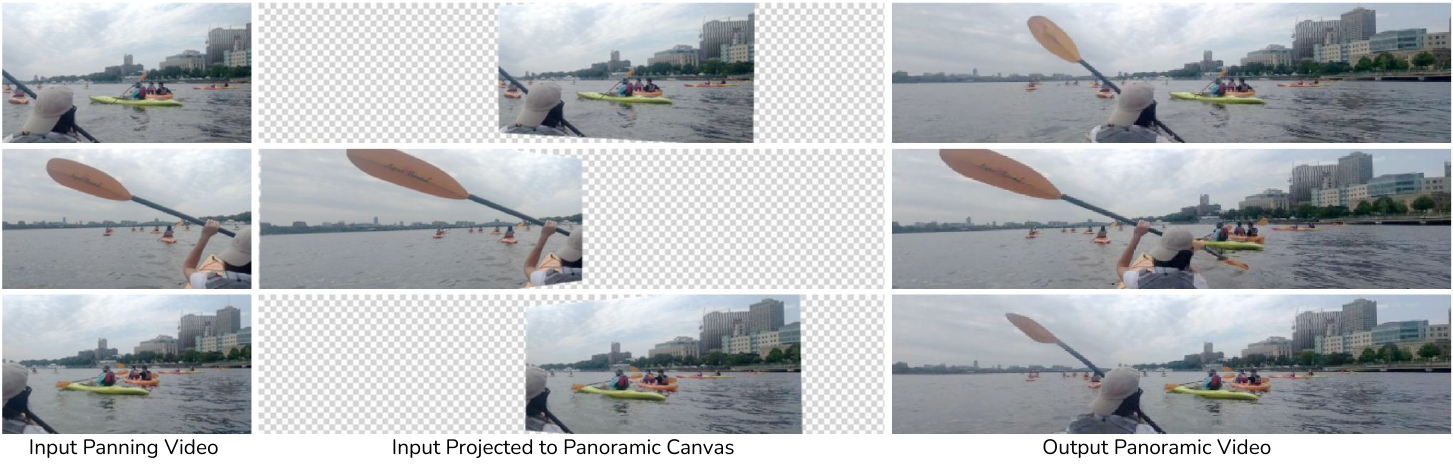}
    \caption{Given a casually-captured panning video, our method synthesizes a coherent panoramic video, depicting the full dynamic scene.  Our framework projects the input video on top of a panoramic canvas and harnesses a generative video model to synthesize realistic and consistent dynamic content in the unknown  regions. Note that the kayaker's paddle moves realistically, even when it is out of frame in the input video.  
    }
    \label{fig:teaser}
\end{teaserfigure}

\keywords{Video panorama, video completion, space-time outpainting, generative video models.}

\ccsdesc[500]{Computing methodologies~Computer vision}

\maketitle

\section{Introduction}
\label{sec:intro}
\tali{The motivation for me for panoramic capture is when there is a scenery that we wish to capture that extents over the field of view of the camera, and we aim to have a single view of it, rather a scattered set of images. Then we can say that this task is well-established for static scenes, but remains open for dynamic ones.}
When visiting a place, for example while traveling, we often want to capture the moment, to help us remember what it was like to be in that place.  Most of us capture a handful of photos with our smartphones for this reason, but the sense of immersion is lost when played back as a sequence of stills -- the sense of the scale of the space is missing.  
We can create a single view that extends beyond the field-of-view of the camera by stitching multiple exposures into a panoramic image. 
Many video cameras can automatically stitch the frames of a panning video into a panoramic still image, so long as the scene is static.
While we can capture a space with an image panorama, however, the experience of the dynamic scene in the moment, filled with moving people, cars, trees, water, etc., is lost.

In this paper, we propose to construct panoramic videos from casually-captured panning video of general dynamic scenes, completing both the space and time spanned by the panoramic video volume.  
We take as input a video that can include not just a single pan, but multiple pans in one capture, \eg, panning from left to right and then back left again.  In this more general multi-pan setting, we have both an opportunity to ground the video with knowledge of “what happened later” as we pan back to a spatial location at a later time, but also the challenge of consistently interpolating across gaps in time to answer the question “what happened in between?”

Our approach is to register the input video frames into a single video volume, leaving space-time regions outside of the input unknown, then complete the unknown volume regions. This task is challenging since in a typical capture, the number of unknown pixels in this volume outnumbers the known pixels, and we cannot assume that the unknown regions are stationary. To solve this problem we need a powerful and realistic prior model of video, and a method to apply this prior to complete the video volume consistently. We demonstrate results with both a diffusion-based model (Lumiere~\cite{bar2024lumiere}) and a token-based model (Phenaki~\cite{villegas2022phenaki}). 
The main technical problem we address is how to constrain and condition these video generation models, which have limited context windows, to complete a panoramic video of arbitrary length and width. To tackle this challenge we apply coarse-to-fine synthesis and spatial aggregation techniques to realistically and consistently complete unknown regions of the video volume.

 

\forrester{rewrote the contributions. still not that happy}
\noindent In summary, our contributions include: 
\begin{itemize}[topsep=3pt,parsep=1ex,leftmargin=*]
\item{The first system for creating video panoramas from general, panning input videos that include moving people and objects.}
\item{Adaptations of the base algorithm for diffusion- and token-based video generation models, and an analysis of their relative strengths and weaknesses. }
\item{A new dataset of video panoramas (cropped from 360-degree videos) with synthetic panning camera motion.}
\end{itemize}

\ignore{

\brian{old intro and comments below}

 We present a method for taking a casually-captured panning video and turning it into a video panorama, as if the original video were captured with a panoramic video camera.

To better capture motion over time, we aim to handle videos that pan over the target scene multiple times. Multiple pans means that the same object, for example a pedestrian, may start out as visible, disappear as the camera pans away, then reappear in a different location as the camera pans back. Objects that repeatedly appear and disappear introduce a long-range \emph{interpolation} problem: the stitched video must smoothly interpolate between each appearance of the object \sz{$\rightarrow$ of each object in the scene?}. This long-range interpolation problem distinguishes our setting from video outpainting (e.g. \cite{yu2023magvit,fan2023hierarchical}, where interpolation across long periods of time is not required. 

Our contributions include:
\begin{itemize}
    \item The first system for creating video panoramas from general input videos including moving people and objects.
    \item A case study on applying a foundation model to a new video generation problem, including lessons for the design of future video generation models.
    \item XXXX
\end{itemize}
\sz{I would add a sentence about why using a wide angle panoramic video camera is not feasible solution. does it require specific equipment, is it expensive? doesn't it exists on regular mobile phones? with our method we can apply on already captured video, etc.}

}
\section{Related Work}
\label{sec:related}

\subsection{Image Panorama Stitching}
A long line of work has focused on the problem of \emph{panorama stitching}~\cite{xiong2010fastpanostitching, brown2007automatic, szeliski2000creating, shum1997panoramic, szeliski2007image, steedly2005efficiently}, \ie, simulating a wider field-of-view image from a set of images captured by a camera rotating in place. These methods typically involve a series of steps including (1) sparse feature-based or semi-dense registration~\cite{lowe2004distinctive}, (2) rigid or depth-compensated alignment, and (3) image blending~\cite{gracias2009fast, perez2023poisson, burt1987laplacian}, to resolve seams and inconsistencies across observations. A major failure mode of image panorama acquisition is in scenes with dynamic objects---since significant scene motion can cause failures in both registration and compositing.

\subsection{Video Panoramas}
The limitation of image stitching in handling moving objects has been explored in a series of work expanding image panoramas to the video domain~\cite{agarwala2005panoramic, couture2011panoramic, rav2007dynamosaicing}. Commonly referred to as panoramic video \emph{textures}, these methods use a graph-cut formulation to solve the panorama blending problem across both spatially and temporally varying observations, and are able to produce panoramas with motion that can even loop seamlessly~\cite{liao2013automated, liao2015fast}.
Still, these methods are restricted to modeling \emph{textural} motion, \eg shaking trees and flowing water, and cannot resolve inconsistencies resulting from transient objects, \eg a person walking across the scene. This is largely due to the fact that these methods do not generate novel observations of the scene---rather, their goal is to consolidate \emph{existing} observations into a consistent representation. 



\subsection{Video Completion}
Video completion focuses on completing missing pixels given observed pixels as context. Many methods retrieve visual features (\eg pixels, patches, segments, templates, proposals) from the observed regions to fill in the missing regions ~\cite{wexler2007space, gao2020flow, liCvpr22vInpainting, hu2020proposal, zhou2023propainter, ilan2015survey, huang2016temporally}. However, these methods tend to fail or produce low-fidelity results when the observation is sparse or incomplete (\eg heavily-masked video, object insertion).

With recent advances in generative models~\cite{bar2024lumiere,   villegas2022phenaki,  ho2022imagen, blattmann2023align, zhou2022magicvideo, guo2023animatediff, blattmann2023stable}, many methods adapt pretrained generative models for the task of video completion~\cite{zhang2023avid, bar2024lumiere} or train versatile generative models on a suite of tasks, where video completion is one of them~\cite{yu2023magvit, kondratyuk2024videopoetlargelanguagemodel}. While the generative completion methods succeed on a broader range of scenarios (\eg foreground/background replacement, transient motion), they fail on the task of panoramic video completion, which requires going beyond the model's temporal and spatial context window, interpolating temporally-distant observations and completing regions under a panning video mask.

\begin{figure*}
    \centering
    \includegraphics[width=\linewidth]{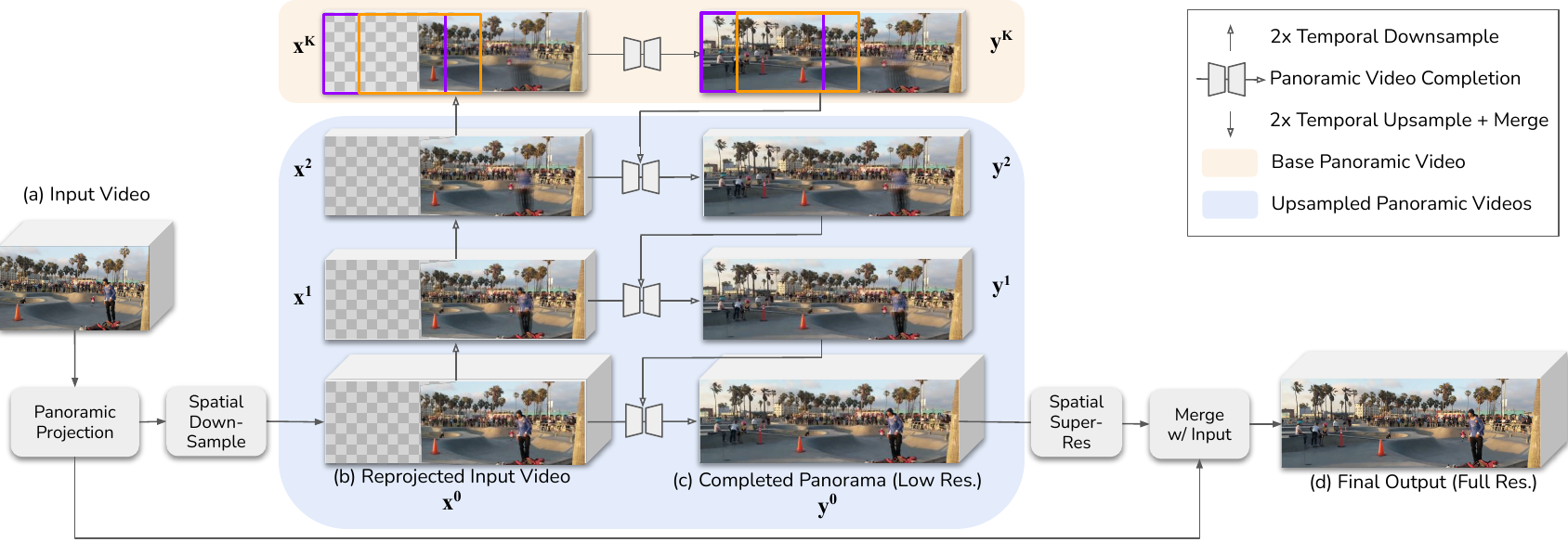}
    \vspace{-0.7em}
    \caption{ Temporal coarse-to-fine. The input video (a) is projected on to a unified panoramic canvas using estimated camera parameters. The reprojected input video (b) is temporally downsampled with temporal prefiltering. A base panoramic video is synthesized at the coarsest temporal scale (top),
then gradually refined by temporal upsampling, merging, and resynthesis (c). Finally, a spatial super-resolution pass is applied and the original input pixels are merged with the result to produce the output video (d).}
    \vspace{-1em}
    \label{fig:system_diagram}
\end{figure*}

\section{Method}
\label{sec:method}

The input to the method is a video captured with a panning camera that sweeps over a scene containing moving people and objects. The output is a complete panoramic video of the same duration as the input, but spatially wide enough to capture the entire sweep of the input camera (e.g., Fig.~\ref{fig:teaser}). 
The output panoramic video should match the input video in the known input regions, and should look realistic and consistent.  Since the extent of the unknown content 
may span a significant portion of the video in both time and space, we harness the power of a generative video prior to synthesize the missing regions.

\tali{Maybe pose it first as a challenge: While existing text-to-video models encode powerful priors about our dynamic world, a pivotal challenge in utilizing them for our task is that they operate on a fixed spatio-temporal context, . It might be useful to point out another challenge of maintaining high-fidelity to the source video in the known regions.}
\tali{Mask-respecting aggregation might be too vague at this point.} 
While existing text-to-video models encode powerful priors about our dynamic world, a pivotal challenge in utilizing them for our task is their limited  spatio-temporal context window. 
We overcome this restriction by adopting a coarse-to-fine approach in the temporal dimension and mask-respecting aggregation in the spatial dimension (Fig.~\ref{fig:system_diagram}). To ensure consistent motion across time, we first temporally downsample the video to the model's context window length and complete a base panoramic video by aggregating the model predictions in sliding spatial windows (Sec.~\ref{sec:spatial}).
We then progressively restore the temporal details by temporal upsampling, merging with the input video, and resynthesizing pixels outside the input regions (Sec.~\ref{sec:coarse-to-fine}). Optionally, the model may be finetuned at test-time to further improve fidelity of the completed video (Sec.~\ref{sec:finetuning}).    
\subsection{Preliminaries: Video Generation Models}

\newcommand{\PIX}{\mathbf{x}}
\newcommand{\TOK}{\mathbf{z}}
\newcommand{\ITOK}{\hat{\mathbf{z}}}
\newcommand{\PMASK}{\mathbf{m}}
\newcommand{\PMASKTOK}{\mathbf{m}_z}
\newcommand{\TMASK}{\mathbf{\mu}}
\newcommand{\POUT}{\mathbf{y}}
\newcommand{\IOUT}{\hat{\mathbf{y}}}
\newcommand{\ENC}{enc}
\newcommand{\DEC}{dec}
\newcommand{\XF}{xf}

To illustrate the generality of the method, we employ two video generation models in our experiments: Lumiere~\cite{bar2024lumiere} and Phenaki~\cite{villegas2022phenaki}. Lumiere is a space-time, pixel-diffusion model with a two-stage cascade: a base model that produces 80 frames of $128\times128$ pixels, followed by an upsampling stage to $1024\times1024$ pixels. Phenaki is a token-based model with an encoder/decoder pair to translate between pixels and the latent token space, as well as a two-stage cascade: first 11 frames of $160\times96$ pixels, then $320\times192$ pixels. 

\subsection{Video Registration and Setup}

In the remaining sections we use the following notation for the intermediate variables:
\begin{center}
\begin{tabular}{ c c }
 \hline
 $\PIX^k$ & input video at temporal scale k\\
 $\PMASK^k$ & mask at temporal scale k\\
 $\IOUT^k_{up}$ & 2x temporal upsampled $\IOUT^k$\\
 $\IOUT^k_{merge}$ & $\IOUT^k_{up}$ merged with $\PIX^k$ \\
 $\POUT^k$ & completed video at temporal scale k\\
 \hline
\end{tabular}
\end{center}

To prepare a video for processing we project the input video onto a panoramic canvas to produce input frames $\PIX^0$ and corresponding mask frames of valid pixels $\PMASK^0$ in the panorama coordinate system. For videos with pure panning camera motion (rotation only), we can use a fast, standard homography solver~\cite{Hartley2004}. When camera parallax is present, we solve for a full 3D camera path using a more expensive, robust SLAM system~\cite{zhang2022structure}. We ignore the translation of the camera and compute the elevation $\theta$ and azimuth $\phi$ of each input pixel's ray relative to the first frame's camera direction, then project each ray to an equirectangular canvas.

\tali{consider explicitly denoting the resolution to it will be clear that the downsampling is only temporally.} \forrester{not sure what you mean here}
We further prepare temporally-downsampled versions of the input $\{\PIX^k\}_{k=1}^K$ such that the coarsest input $\PIX^K$ fits exactly in the model's context window (80 frames for Lumiere, 11 for Phenaki). To avoid temporal aliasing we apply simple temporal prefiltering with a box blur before subsampling from $\PIX^0$.

\begin{figure*}
    \centering
    \includegraphics[width=1\linewidth]{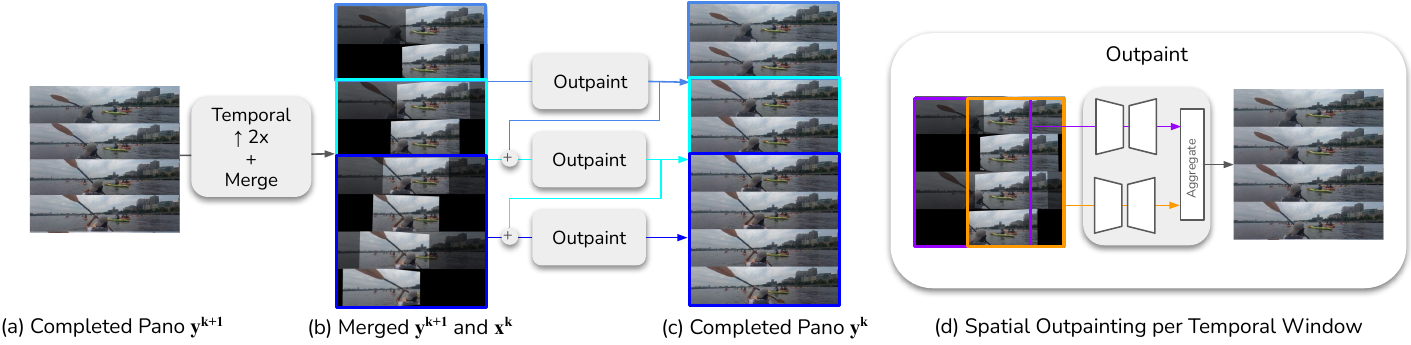}
    \caption{Upsampling and outpainting. The completed panorama from the previous level $\mathbf{y}^{k+1}$ (a) is temporally-upsampled and composited with the current level input video $\mathbf{x}^k$ to form a partially-completed input $\hat{\mathbf{y}}^k_{merge}$ (b, input pixels shown highlighted). The model uses the full $\hat{\mathbf{y}}^k_{merge}$ for context and resynthesizes content outside the input mask to complete the next level panorama $\mathbf{y}^{k}$ (c). In the time dimension, the model is applied in a sliding-window fashion with half-window overlap. In the spatial dimension, multiple overlapping predictions are computed in parallel, then aggregated and a sample is drawn from the average (d).}
    \vspace{-1em}
    \label{fig:merge_diagram}
\end{figure*}

\subsection{Base Panoramic Video Completion}
\label{sec:spatial}

The first step is to complete a panoramic video at the coarsest temporal resolution by spatial outpainting.
The input video $\PIX^K$ is in general wider than the model's native aspect ratio. We downscale $\PIX^K$ to match the model's native height and use multiple, overlapping spatial windows to span the panorama width. The distributions predicted by the model in each window are averaged, then a new sample is drawn from the average. This approach applies to both diffusion and token-based models, as explained below.

\subsubsection{Diffusion}
\label{sec:diffusion_spatial}
For a diffusion model, averaging overlapping windows is a form of MultiDiffusion~\cite{bar2023multidiffusion}. We crop the projected panoramic canvas to each input window, then outpaint any regions outside the valid pixels $\PMASK$ using the mask-conditioned version of Lumiere. The $\mu$ and $\Sigma$ predictions are averaged, then a new sample is drawn using DDPM~\cite{ho2020denoising}. We found that the shape and motion of the panorama masks caused boundary artifacts with the original mask-conditioned model, so we finetuned the model on a dataset of natural videos masked by synthetic panorama masks designed to mimic real $\PMASK$. \erika{do we have to say something about the multidiffusion window size here? using 256 vs. 128. can also be in supp}

\subsubsection{Token-based}
\label{sec:token_based_spatial}
\begin{figure}
    \centering
    \includegraphics[width=1\linewidth]{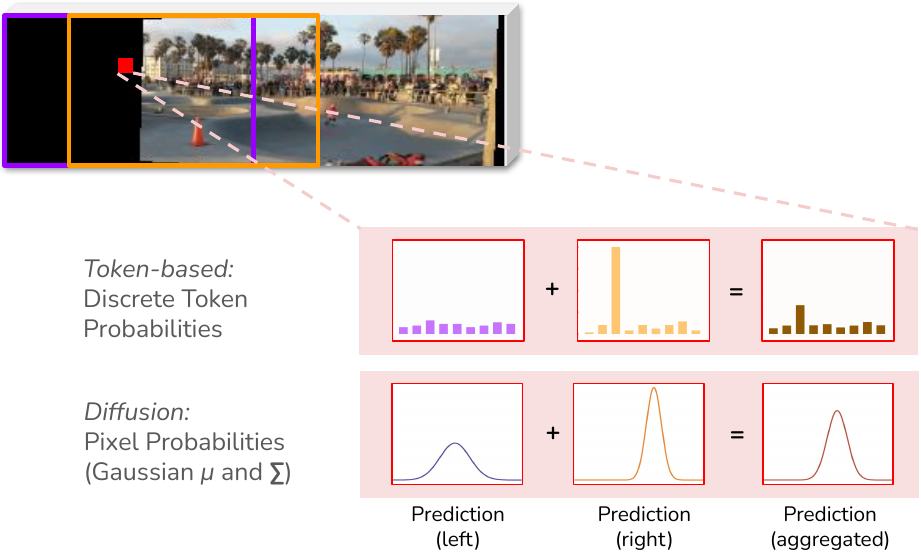}
    \caption{Spatial aggregation of predicted distributions. To generate a sample in the overlap (red), we linearly interpolate the two predicted probability distributions (purple, orange) and sample from the aggregated distribution (brown). With a token-based method the distribution is a discrete distribution over the vocabulary. With diffusion, the distribution is a Gaussian distribution over pixel values, represented by $\mu$ and $\Sigma$.}
    \vspace{-1em}
    \label{fig:spatial-aggregation}
\end{figure}

For a token-based model like Phenaki, spatial aggregation can be performed by averaging the predicted probability distributions over the tokens before sampling. Fig.~\ref{fig:spatial-aggregation} illustrates spatial aggregation with a simplified case of two overlapping spatial windows. The red patch represents one of the tokens to be generated and lies within both the left (purple) and the right (orange) windows. Each window of masked tokens can be input to the transformer network to predict a probability distribution for the red token. These distributions are then averaged and a predicted token is drawn from the averaged distribution.

Note that Phenaki employs causal masking during training, which means the model cannot complete earlier frames based on later frames at test time. To work around this issue, the base panorama $\PIX^K$ is completed in two passes, one forward and one backward (time-reversed), and the valid pixels of the forward and backward passes are merged (please see supplemental material for details). Since later temporal upsampling steps always start from a complete panorama, this approach is only necessary for $\PIX^K$. 



\subsection{Temporal Coarse-to-fine}
\label{sec:coarse-to-fine}

To restore the original temporal dimension of the video, we progressively complete panoramic videos at upsampled temporal resolutions. For each level $k\in[K-1,0]$, we take the input video $\PIX^k$ and the completed, coarser-level panoramic video $\POUT^{k+1}$ and combine them to produce a complete panoramic video $\POUT^k$ (Fig.~\ref{fig:merge_diagram}). Intuitively, we want the video generation model to provide temporal details for $\POUT^k$ that are consistent with the input pixels in $\PIX^k$ and the coarse-level context in $\POUT^{k+1}$. To achieve this, we apply three steps: (1) temporal upsampling, (2) merging with the input pixels, and (3) resynthesis.

We first temporally upsample $\POUT^{k+1}$ to create a $\IOUT^k_{up}$ that is frame-rate matched, and composite $\PIX^k$ over $\IOUT^k_{up}$ to form a merged $\IOUT^k_{merge}$. Optionally, we may align $\PIX^k$ to $\IOUT^k_{up}$ prior to compositing using grid-warp-based optical flow~\cite{Szeliski2010} and color histogram matching. We found spatial and color alignment useful when using Phenaki, but unnecessary when using Lumiere. The resulting $\IOUT^k_{merge}$ matches the input video inside $\PMASK^k$ but lacks temporal details outside. 

We adapt the resynthesis algorithm to the base model type, as follows:

\subsubsection{Diffusion.} 
Resynthesis using a diffusion model is controlled by the  mask conditioning signal. A full-frame mask is applied to the odd-numbered frames of $\IOUT^k_{merge}$, and the original mask $\PMASK^k$ to the even-numbered frames. That is, we constrain the diffusion sampling to maintain the temporally-upsampled, generated pixels at the odd-numbered frames, and allow resynthesis of the generated pixels only at the even-numbered frames. For videos with fast motion (e.g., the kayak paddle in Fig.~\ref{fig:teaser}), the temporally-upsampled pixels at odd-number frames may also need to be resynthesized. In this case we maintain the full-frame masks at odd frames for the first $1/8$ of the sampling schedule, then switch to the input mask $\PMASK^k$ for the entire window to allow the model to synthesize new temporal details. Resynthesis is applied over the entire sequence using sliding temporal windows with an overlap of half the window length (40 frames for Lumiere). 

\subsubsection{Token-based.}

With a token-based model, the pixels outside $\PMASK^k$ are resynthesized by masking and regenerating the corresponding tokens:
\[
\ITOK^k_{merge} = \ENC(\IOUT^k_{merge})
\]
\[
\TOK^k = \XF(\ITOK^k_{merge}\odot \PMASKTOK^k),
\]
where $\ITOK^k_{merge}$ is the set of tokens encoded from $\IOUT^k_{merge}$, $\PMASKTOK^k$ is the token-level mask, $\TOK^k$ is the set of resynthesized tokens, $\ENC$ is the token encoder, and $\XF$ is the token transformer network. The full sequence $\POUT^k = \DEC(\TOK^k)$ is constructed using sliding temporal windows with an overlap of half the window length (5 frames for Phenaki, as in ~\cite{villegas2022phenaki}).

\subsection{Inference-Time Finetuning}
\label{sec:finetuning}

Some modules of the video generation model can be optionally finetuned at inference time to further improve results. The Phenaki model's encoder/decoder architecture in particular incurs some fidelity loss, so the base Phenaki model cannot exactly reproduce the original video pixels. To better align the result with the input video, we finetune the Phenaki decoder $\DEC$ on patches of valid pixels $\PIX\odot \PMASK$ prior to synthesizing the final result. Finetuning helps to preserve details of the input in the outpainted regions. Lumiere is a pixel-diffusion model without an encoder and decoder step, and we found it produced results closer to the input video overall, though high-frequency details could likely be improved with finetuning (see Sec.~\ref{sec:discussion_and_limitation}).

\section{Results}

We evaluate our approach in two settings: real-world casual panning captures, and ``synthetic'' panning videos generated by applying a moving crop window to videos captured with a 360-degree camera. The synthetic setting allows us to compare the models' performance directly against a ground-truth panoramic video.

\subsection{Baseline Methods}

\begin{figure*}
    \centering
    \includegraphics[width=0.98\linewidth]{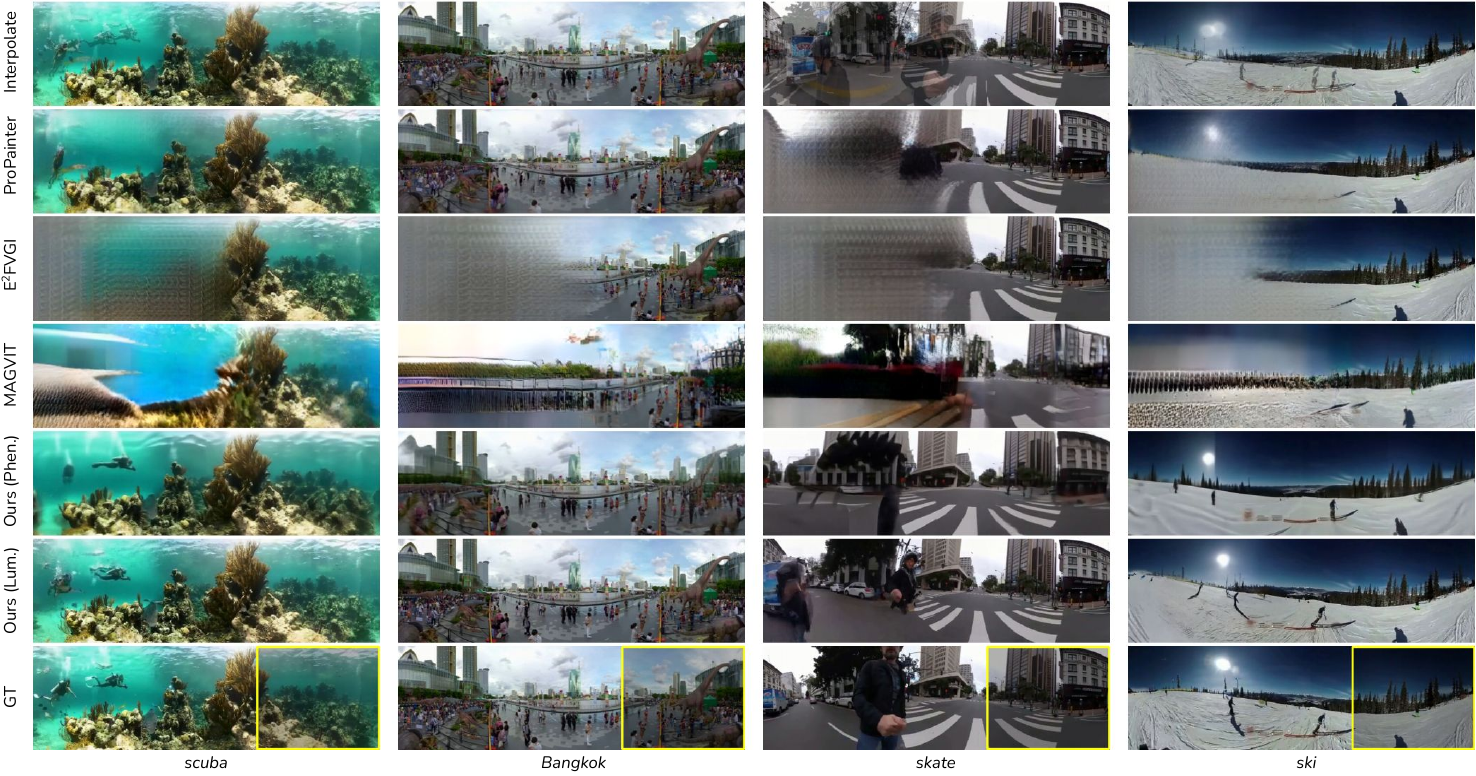}
    \vspace{-0.8em}
    \caption{Comparison with baseline methods. From top to bottom: linear interpolation between pixels based on time produces sharp results for stationary regions, but does not interpolate motion. ProPainter~\cite{zhou2023propainter} and E$^2$FGVI~\cite{liCvpr22vInpainting} are flow-based methods that can produce realistic results in stationary regions (scuba, Bangkok), but fail for moving cameras (skate, ski) or moving objects away from the input window (divers on left in scuba). MAGVIT~\cite{yu2023magvit} is a video-generation method but does not generate on a common panorama canvas, so it loses information away from the input window. Our results use a coarse-to-fine approach to build a consistent panoramic video and better match the ground-truth. Bottom: ground truth video with input window marked in yellow. See supplemental material for video results. }
    \vspace{-0.8em}
    \label{fig:comparison_results}
\end{figure*}

We include four baseline methods for comparison (Fig.~\ref{fig:comparison_results}): a simple linear interpolation baseline, two flow-based video inpainting algorithms, and a recent video generation model MAGVIT~\cite{yu2023magvit} that demonstrated video panorama outpainting. 

\paragraph{Linear interpolation baseline.} For the linear baseline, the output color at pixel $\mathbf{p}$ is a linear interpolation of the closest before and after frames, or the nearest neighbor frame if only before or after exists. The result matches the input video exactly and matches any stationary elements in the scene. Moving objects exhibit the expected ghosting artifacts, with extreme failures in the case of non-stationary camera.

\paragraph{Flow-based baselines.} We include two methods for video inpainting using optical flow: ProPainter~\cite{zhou2023propainter} and E$^2$FGVI\cite{liCvpr22vInpainting}. These methods are tasked with completing the region outside the valid pixels $\PMASK$. Both methods internally estimate flow, so the only inputs are $\PIX^0$ and $\PMASK^0$.

\paragraph{MAGVIT baseline.} For the MAGVIT baseline, we apply repeated horizontal outpainting to extend the input video to outside the panorama canvas, then crop. Due to the temporal window of MAGVIT being limited to 16 frames, we subsample a portion of each video to obtain a single-direction pan of the full scene from left to right, and evaluate only on this subset of frames.

\subsection{Quantitative Evaluation}
Besides typical similarity metrics (PSNR, LPIPS~\cite{zhang2018perceptual}) and single-video FID~\cite{9423034}, we also compute optical flow EPE (endpoint error) to measure the consistency of the generated motions. We compute flow between consecutive frames of the groundtruth and the output video and measure their L2 difference. When evaluating small motions at low image resolution, we empirically found grid-based flow~\cite{Szeliski2010} to produce more reliable sub-pixel alignments than network-based flow (e.g. RAFT~\cite{teed2020raft}). For pixel-level metrics (PSNR, EPE), we separately evaluate static and dynamic regions (see supplemental for details).

\subsection{Synthetic Panning Videos}
\label{sec:synthetic_results}

To create synthetic panning videos we center crop an input video captured with a 360-degree video camera and add a moving crop window over 88 frames at 15fps  (Figure~\ref{fig:synthetic_results}). We then apply our system to complete a new video panorama. We curated 12 360-degree videos licensed under Creative Commons and plan to include these videos along with our results upon publication. 

Synthetic videos allow us to evaluate the model's output directly against a ground-truth video panorama. Note that our goal is to generate a plausible completed panorama, not recreate an input panorama. However, our model should recreate stationary scene elements as closely as possible. 

Fig.~\ref{fig:synthetic_results} shows results on two stationary camera videos (``scuba'', ``Bangkok'') and two moving camera videos (``skate'', ``ski''). Our method recovers static regions faithfully, and renders moving objects in plausible positions (diver in ``scuba'', skiers in ``ski'') even under challenging moving-camera settings. Certain scenarios prove too difficult to resolve: for example, the person in ``skate'' is observed very briefly but undergoes large motion. The model realistically renders the missing person in the first frame, but it struggles to complete the middle frame. 

We show qualitative comparisons with baselines in Fig.~\ref{fig:comparison_results}. Linear interpolation produces obvious ghosting artifacts for moving objects (the diver in ``scuba'', skier in ``ski''), and fails completely for camera motion (``skate''). The MAGVIT prediction is reasonable near the observed region, but quickly degrades the greater the distance from the input window.


Quantitative results are shown in Table~\ref{tab:synthetic_results}. For all results on ``Ours'', we generate four samples and manually select the best. Our diffusion-based method performs the best across all metrics except the static split of PSNR and EPE. The baseline interpolation method performs best on the static splits, which is expected given that it perfectly reproduces stationary parts of the synthetic videos. However, the interpolation method performs poorly on EPE dynamic split (1.92 vs. 1.67 and 1.25 for our token-based and diffusion-based methods, respectively).


The flow-based methods ProPainter~\cite{zhou2023propainter} and E$^2$FGVI~\cite{liCvpr22vInpainting} both assume small mask regions and large frame-to-frame overlap. Nevertheless, ProPainter~\cite{zhou2023propainter} handles videos with stationary camera (e.g. scuba, Bangkok) surprisingly well, even producing a higher PSNR than our Phenaki-based result. Subjective quality is lower, however, especially for scenes with a moving camera (Fig.~\ref{fig:comparison_results}, skate, ski).

The video models allow our method to produce more realistic motion than baselines in the inpainted regions, with a static/dynamic EPE of 0.05/1.25 (Lumiere) and 0.07/1.67 (Phenaki) compared to 0.12/1.70 for ProPainter.


\begin{table}
    \centering
    \caption{Quantitative results on synthetic panning videos, computed on the inpainted regions (further split into static and dynamic regions for pixel-level metrics). MAGVIT* is evaluated on a subset of frames (Sec.~\ref{sec:synthetic_results}).}
    
    \begin{tabular}{c|c|c|c|c|c|c}
        Method & \multicolumn{2}{c|}{PSNR $\uparrow$} & LPIPS $\downarrow$ & VFID $\downarrow$ & \multicolumn{2}{c}{EPE $\downarrow$} \\
        & sta & dyn & & & sta & dyn \\
        \hline
        Interpolate & \textbf{29.4} & 19.1 & 0.10 & 0.09 & \textbf{0.04} & 1.92 \\
       ProPainter  & 24.7 & 19.6 & 0.19 & 0.21 & 0.12 & 1.70\\
       E\textsuperscript{2}FGVI  & 18.2 & 16.6 & 0.36 & 0.47 & 0.63 & 2.03 \\
       MAGVIT*  & 12.9 & 12.4 & 0.41 & 0.57 & 1.17 & 1.92\\
       naive Phenaki  & 18.3 & 16.4 & 0.23 & 0.26 & 0.41 & 1.94\\
       Ours (Phenaki)  & 23.2 & 18.4 & 0.20 & 0.19 & 0.07 & 1.67\\
       naive Lumiere  & 18.5 & 18.3 & 0.24 & 0.18 & 0.41 & 1.94\\
       Ours (Lumiere)  & 28.5 & \textbf{20.8} & \textbf{0.09} & \textbf{0.05} & 0.05 & \textbf{1.25}\\
    \end{tabular}
    \label{tab:synthetic_results}
    \vspace{-2em}
\end{table}
\begin{figure}
    \centering
    \includegraphics[width=1\linewidth]{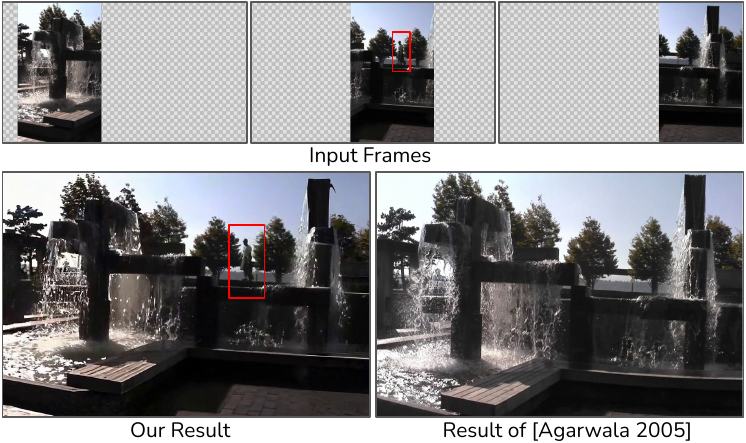}
    \vspace{-1.5em}
    \caption{Comparison with Panoramic Video Textures~\cite{agarwala2005panoramic}. PVT uses a graph-cut formulation to create a looping panoramic video. Our method can create similar videos, but can also include non-stationary features like the person walking behind the waterfall (boxed).}
    \vspace{-1.5em}
    \label{fig:waterfall_results}
\end{figure}

\begin{figure*}
    \centering
    \includegraphics[width=1\linewidth]{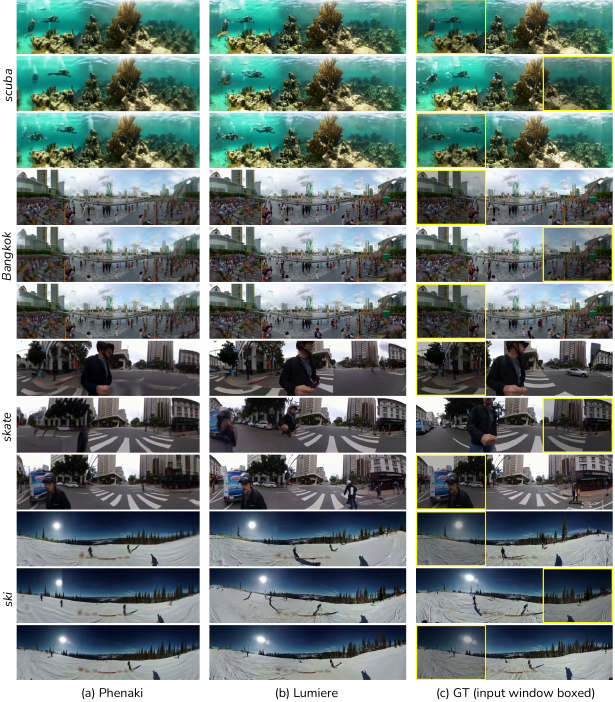}
    \caption{Results on synthetic panning videos. Left: Phenaki model results. Middle: Lumiere model results. Right: ground-truth panoramic video captured with wide-angle camera. Darkened boxed area is the input window shown to the model. Please see supplemental material for full video results.}
    \label{fig:synthetic_results}
\end{figure*}
\begin{figure*}
    \centering
    \includegraphics[width=1\linewidth]{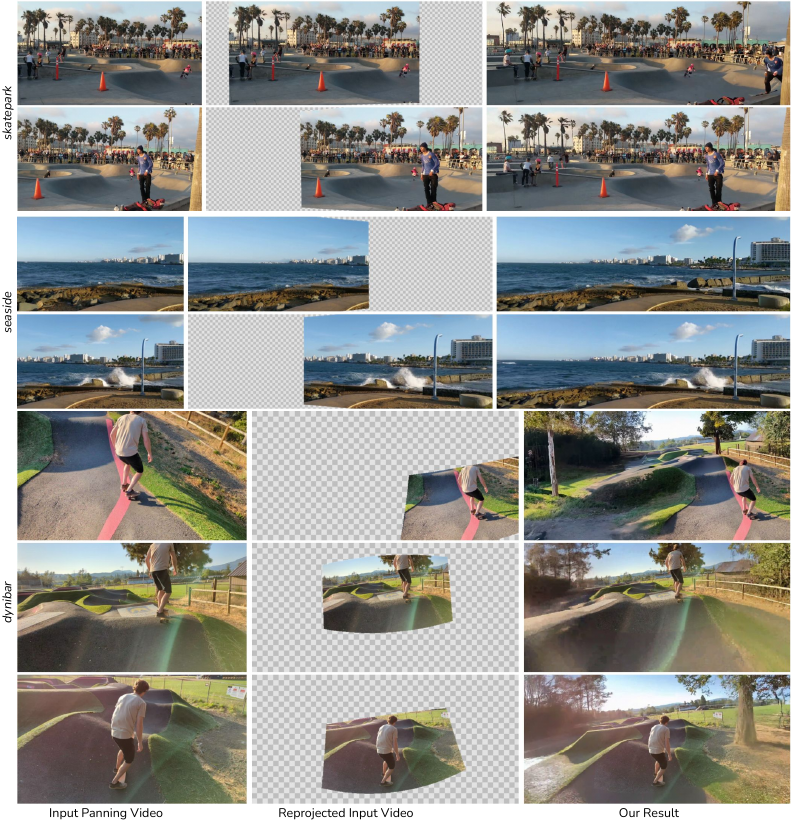}
    \caption{Results on real videos. Left: representative input frames. Middle: frames projected to panorama canvas. Right: our result. Our method synthesizes realistic motions for an unseen person entering the frame (top), ocean waves (middle), and for scenery around a moving camera (bottom). See supplemental material for videos. \todo{Should center column titles with respect to the last example? Make figure 8/9 title font size consistent.}}
    \label{fig:real_results}
\end{figure*}

\subsection{Real Panning Videos}

To further evaluate performance on real-world panning captures, we captured a set of 10 panning videos using a phone video camera. Three of these videos contain panning in both directions, while the rest pan in a single direction. Additionally, 6 of the 10 videos are filmed with a vertical aspect ratio. These videos have roughly stationary cameras that we stabilize onto the panoramic canvas using homographies~\cite{Hartley2004}. Fig.~\ref{fig:real_results} shows representative examples of real videos. We do not have ground-truth panoramic videos to compute quantitative results against, but we observe that overall quality is similar to the synthetic panning results. We additionally process and compare the waterfall video from the original panoramic video textures work~\cite{agarwala2005panoramic}  (Fig.~\ref{fig:waterfall_results}).

\subsection{Ablations}

Besides the alternative baselines, we analyze two main ablations of our diffusion-based method: 1) ``naive'' Lumiere without mask finetuning, and 2) our method with temporal MultiDiffusion instead of temporal coarse-to-fine. Ablation results are shown in Fig.~\ref{fig:ablation_results} and Fig.~\ref{fig:multidiffusion_ablation}.
Additional ablations for our token-based method can be found in supplemental.

\paragraph{Naive Lumiere.} We use the baseline mask-conditioned Lumiere model from~\cite{bar2024lumiere} along with spatio-temporal MultiDiffusion to complete the panorama videos. Since the mask-conditioned model was trained on static masks, it produces significant artifacts on the dynamic-mask panorama videos, including visible seams around the mask boundary, color issues, and large blobs (Fig.~\ref{fig:ablation_results}). Quantitative comparisons are shown in Table~\ref{tab:synthetic_results}.

\paragraph{Removing temporal coarse-to-fine.} We ablate the temporal coarse-to-fine component and replace it with temporal MultiDiffusion with window sizes of 80 frames and stride of 40 frames. A comparison is shown in Fig.~\ref{fig:multidiffusion_ablation}. As the camera pans from left to right, the temporal MultiDiffusion result suffers from drift and is unable to propagate the appearance of the person from the earlier frames to the later frames (orange box). Temporal coarse-to-fine generates a more plausible continuation of the person's appearance and motion due to an initial round of coarse completion where the model sees the full extent of the scene within a single temporal window.

\subsection{Computational Cost}
For the Lumiere model, base-resolution inference on a Google Cloud TPU v5p-4 configuration for a 84x128x512 size video (39 Lumiere forward passes, 256 ddpm steps per pass) takes 300 minutes. The super-resolution stage runs diffusion at 8x the base resolution and takes 48 minutes (7 Lumiere super-res forward passes, 32 ddim steps per pass). The one-time fine-tuning of the original Lumiere checkpoint on panoramic masks takes 30 hours on a batch size of 128 for 35K steps.
For Phenaki, inference for a 172$\times$320$\times$96 video (31 Phenaki forward runs) on a TPU v3-8 is 20.4 min; finetuning the Phenaki decoder $\DEC$ takes $\sim$25 min.

Inference for both models can be greatly optimized by parallelizing the $\mathbf{M}$ spatial windows, reducing the runtime to its  $\frac{1}{\mathbf{M}}$.

\section{Discussion and Limitations}
\label{sec:discussion_and_limitation}

The method presented in this paper can complete a panoramic video from a casually-captured panning video. This task would be impossible without a strong prior on realistic videos and motion, which has only recently become available in the form of generative models of video. While panoramic videos from stationary (rotation-only), panning cameras have been shown in limited settings, such as Panoramic Video Textures~\cite{agarwala2005panoramic}, constructing panoramic videos containing large object motions (Fig.~\ref{fig:teaser}) or the entirely shifting visual field of a \emph{moving} camera (Fig.~\ref{fig:real_results}, ``dynibar'') are capabilities unlocked only with a generative video prior. 

\noindent Given the nascent capabilities of generative video models, however, our method has some limitations that could lead to future work: 

\begin{figure}
    \centering
    \includegraphics[width=0.95\linewidth]{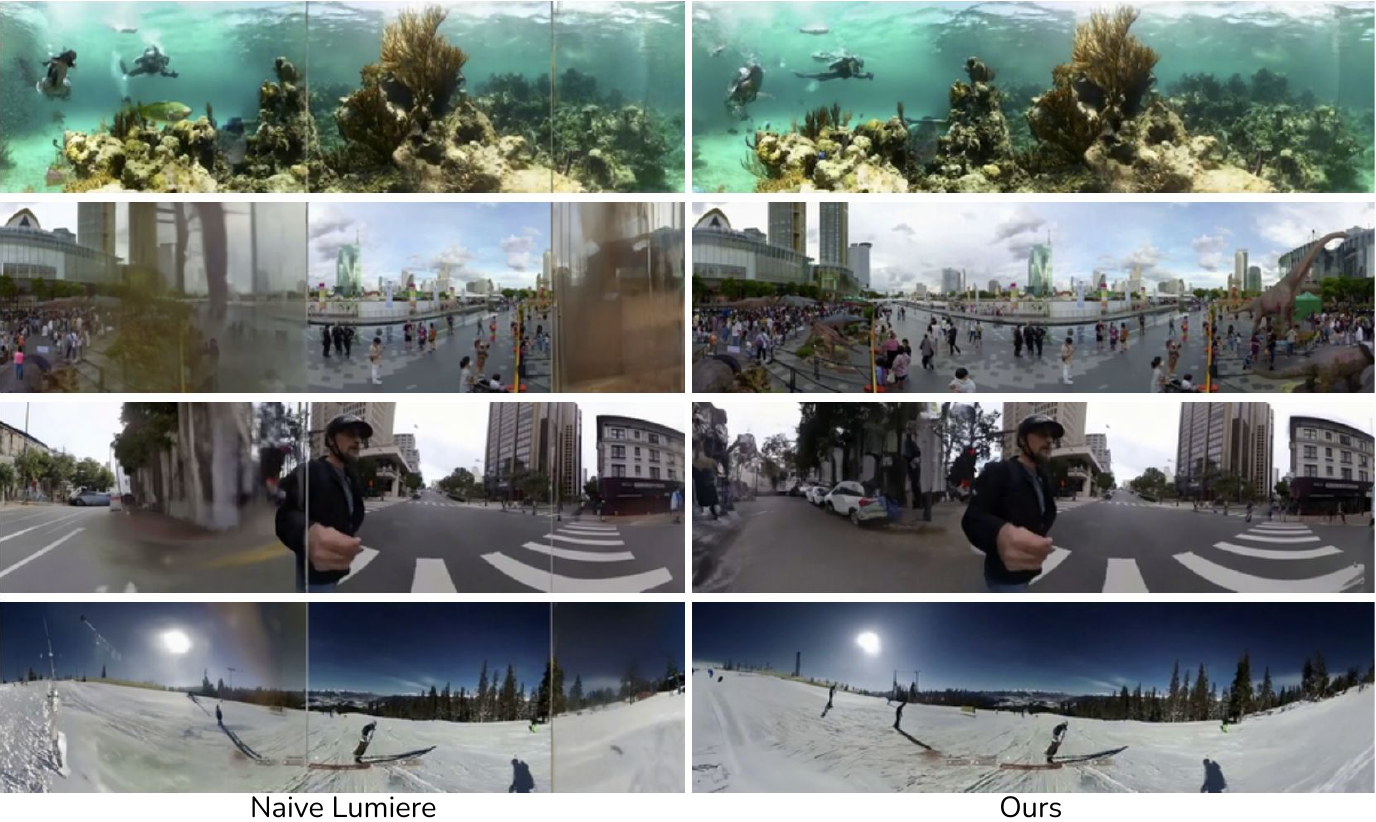}
    \vspace{-1em}
    \caption{Naive Lumiere vs. Ours. Left: Lumiere without panorama mask finetuning or temporal coarse-to-fine. Right: our result. Compare with our full method and ground-truth in Fig.~\ref{fig:comparison_results}.}
    \vspace{-0.9em}
    \label{fig:ablation_results}
\end{figure}

\begin{figure}
    \centering
    \includegraphics[width=0.95\linewidth]{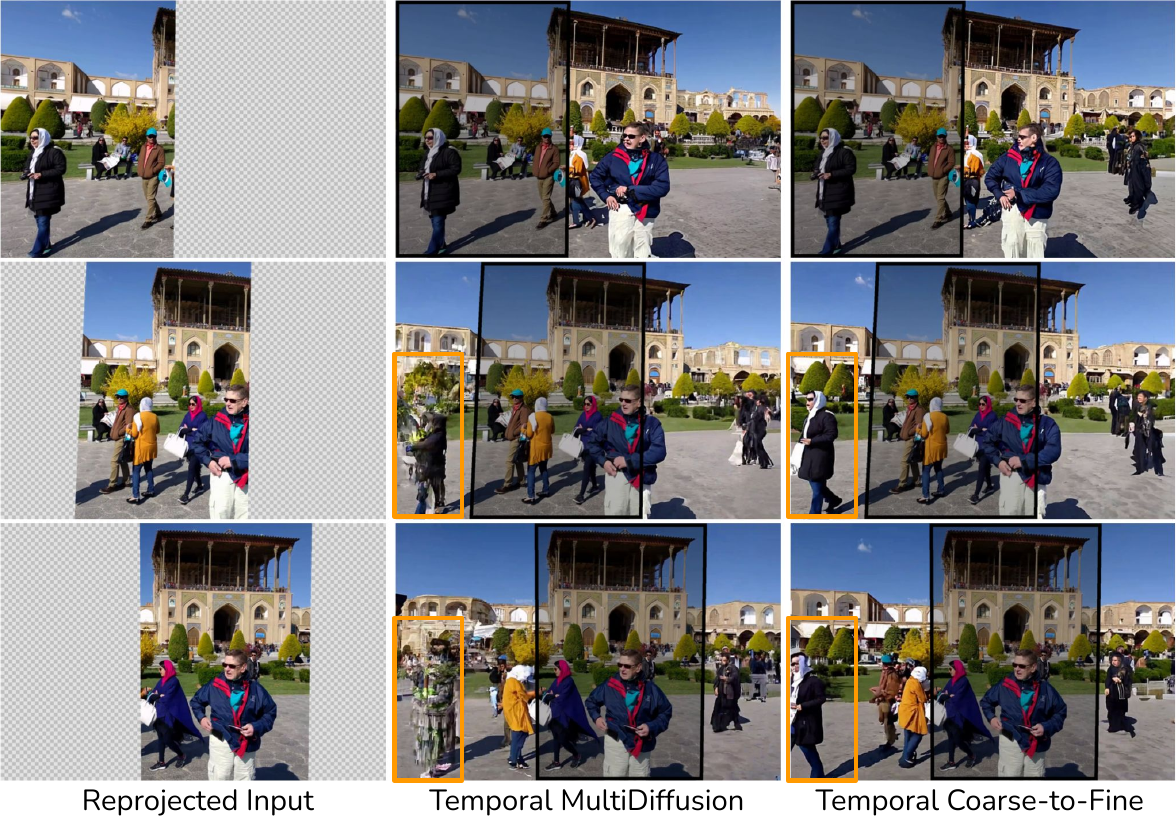}
    \vspace{-1em}
    \caption{Ablation of Temporal Coarse-to-Fine. Coarse-to-Fine synthesis (right) generates more consistent results over long videos than temporal MultiDiffusion (middle). With temporal MultiDiffusion, later generations can drift from the input pixels (orange box), while coarse-to-fine generates a plausible continuation of the pedestrian. Input pixels shown darkened.}
    \vspace{-1.5em}
\label{fig:multidiffusion_ablation}
\end{figure}

\paragraph{Limited context window.} Current video generation models process a limited number of frames simultaneously (80 for Lumiere, 11 for Phenaki). This limited context window necessitates a temporal coarse-to-fine approach  to allow the model to fill in the entire panoramic video consistently (Fig.~\ref{fig:multidiffusion_ablation}). A limited amount of temporal coarsening is possible, however, without completely blurring out fast motion or dropping small objects. For our experiments we used up to 5 levels of temporal coarsening for Phenaki and 2 for Lumiere, for a maximum video length of 172 or 160 frames.

\paragraph{Synthesis quality.} While the quality of the generated video is often convincing, neither model we tested consistently generates photorealistic results. Limitations on synthesis quality are especially noticeable for close-up human faces (e.g., Fig.~\ref{fig:multidiffusion_ablation}, "palace"). 
Our diffusion-based results could potentially preserve the high frequencies of the input videos better by finetuning the Lumiere super-resolution module with mask-conditioning. However, this modification would require adding a new conditioning input and computationally-intensive retraining, and was not done as part of this work. We expect this limitation to be removed in the future.


\paragraph{Latent video diffusion models.} Several recent video generation models~\cite{guo2023animatediff, blattmann2023stable} use a latent diffusion model (LDM) as a backbone to reduce runtime cost. Since the latent space is lower resolution than the image space, applying our method to LDM-based models would likely require careful handling of masking in the encoder and the diffusion model, possibly through a combination of panoramic-mask finetuning and separate masking in the encoder and diffusion model. We leave this exploration for future work.





{
    \small
    \bibliographystyle{ACM-Reference-Format}

}
\appendix

\twocolumn[
    \noindent
    {\sffamily \huge Supplementary Material}
    \vspace{0.5cm}
]


\renewcommand{\thesection}{S-\arabic{section}}


\section{Method Details}

\begin{center}
\begin{tabular}{ c c }
 Variable & Description \\ 
 \hline
 $\PIX^k$ & input video at temporal scale k\\
 $\PMASK^k$ & mask at temporal scale k\\
 $\IOUT^k$ & completed video at temporal scale k\\
 $\IOUT^k_{up}$ & 2x temporal upsampled $\IOUT^k$\\
 $\mathbf{N}^k$ & number of frames at temporal scale k\\
 $enc$ & VQ video encoder\\
 $dec$ & VQ video decoder
\end{tabular}
\end{center}

\subsection{Panorama Registration}
\label{sec:registration}
We use simple homography-based registration to stabilize our real videos. Thus the resulting videos may contain imperfections from stabilization that we can adjust for by applying a coarse warp at various stages of our pipeline, which we describe in Sec. ~\ref{sec:warp}.

\subsection{Panorama Completion}
\label{sec:baselevel}

\begin{figure}[h]
    \centering
    \includegraphics[width=0.98\linewidth]{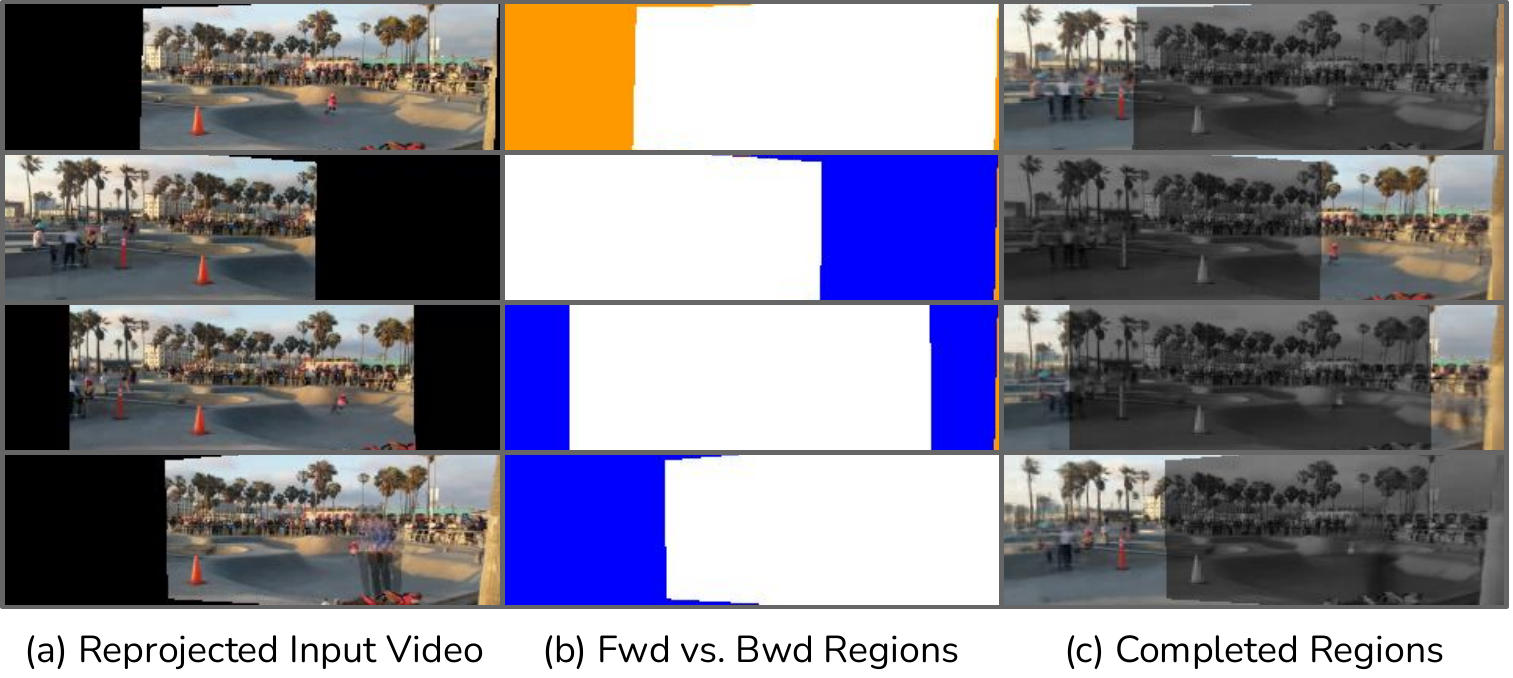}
    \vspace{-1.1em}
    \caption{Base level completion. Given a reprojected input video (a), we run the video generation model at the coarsest level forwards and backwards in time (b), with the forward pass used if an earlier frame contains data at that pixel (blue) and the backwards pass where only the later frames contain data (orange). We combine the two passes to get the completed regions (c).}
    \vspace{-1.1em}
    \label{fig:fwd_bwd_diagram}
\end{figure}

\textbf{Forward-backward pass (Phenaki only, base level).} Since its original application is to extend a given video, Phenaki employs causal masking: the attention layers of $\ENC$ are masked such that later frames attend to earlier frames in the window, but earlier frames are blocked from attending to later frames. This masking causes artifacts when we synthesize regions that are not seen until later frames. To complete those regions (Fig.~\ref{fig:fwd_bwd_diagram}), we run Phenaki both forward and backward over the coarsest level clip $\PIX^K$, taking the result of the forward pass in regions seen previously by the camera (blue) and combining it with the backward pass in the rest of the regions (orange). The result is an 11-frame completed panoramic video $\POUT^K$.

\textbf{Temporal box filtering (Phenaki only).} At each temporal level $k$, we apply a box blur on the full framerate input $\PIX^0$ ($\mathbf{N}^0$ total frames) before frame subsampling. The box filter size for temporal level $k$ is $\mathbf{N}^0/\mathbf{N}^k$ and the temporal stride is $\mathbf{N}^0/\mathbf{N}^k$, both rounded to the nearest integer. We then subsample 1 frame from the center of each temporal window ($\mathbf{N}^k$ total) to obtain $\mathbf{N}^k$ frames for the respective temporal level.

\textbf{Spatial windows.} We downsample the input to fit the video model's height dimension, and span the width dimension with multiple overlapping windows. We use a stride of 32 pixels for Lumiere and a stride of 80 pixels for Phenaki.

\subsection{Spatial/Color Alignment}
\label{sec:warp}

In our temporal coarse-to-fine pipeline (Sec. 3.4), we complete a base panoramic video followed by multiple temporally-upsampled panoramic videos. An important subtlety for the upsampled token-based video completion is that since $\IOUT^k_{up}$ is the output of video generation, spatial details and color may not align exactly between $\IOUT^k_{up}$ and $\PIX^k$ (this mis-alignment is particularly worse for real videos with imperfect stabilization; see Sec. ~\ref{sec:registration}). We found improved results with our token-based method by aligning $\PIX^k$ to $\IOUT^k_{up}$ before merging. We align $\PIX^k$ to $\IOUT^k_{up}$ spatially by computing a coarse flow field, obtaining $\PIX^k_{warp}$. We then perform an adjustment in color space by computing a Gaussian pyramid for  $\IOUT^k_{up}$ and $\PIX^k_{warp}$ and constructing a color-aligned $\PIX^k_{aligned}$ using the 2 finest pyramid levels of $\PIX^k_{warp}$ and the coarsest $n-2$ levels of  $\IOUT^k_{up}$.   The final merged video is computed as $\IOUT^k_{merge} = over(\PIX^k_{aligned}, \IOUT^k_{up})$, where $over()$ is the conventional over-compositing operation~\cite{Porter1984CompositingDI}. 

\subsection{Finetuning Lumiere on Dynamic Masks}
We finetune the mask-conditioned Lumiere model on a dataset of 5 million videos of dimension 128x128, and on masks generated by randomly taking 128x128 crops and augmenting our set of synthetic and real panning video masks. Initializing from the original Lumiere inpainting model weights, we finetune for 35k steps, using a batch size of 128 and the Adafactor optimizer \cite{shazeer2018adafactor} with $\beta_1 = 0.9$ and $\beta_2 = 0.999$. We use a constant learning rate of $1 \times 10^{-5}$. The finetuning continues to optimize for the diffusion denoising objective (squared error loss), with our dynamic masks.

\subsection{Phenaki Decoder Finetuning}
As described in Sec. 3.5, we finetune the decoder $dec_{\theta}$ to restore video details lost during the tokenization process. We finetune one model on all temporal scales, using a batch size of 24 and the Adam optimizer ~\cite{kingma2017adam} with $\beta_1 = 0.9$ and $\beta_2 = 0.9$. We use an initial learning rate of 
$1 \times 10^{-4}$ and decay to a final learning rate of $1 \times 10^{-6}$ over 5000 steps with a cosine schedule. The finetuning objective is: 
\[
\mathcal{L} = \sum_{x \in D}||(dec_{\theta}(enc(x))-x)\odot m||_{2}^{2} 
\] where $D$ is the set of 11-frame clips sampled from different temporal scales in each batch.

\begin{figure*}[h]
    \centering
    \includegraphics[width=0.95\linewidth]{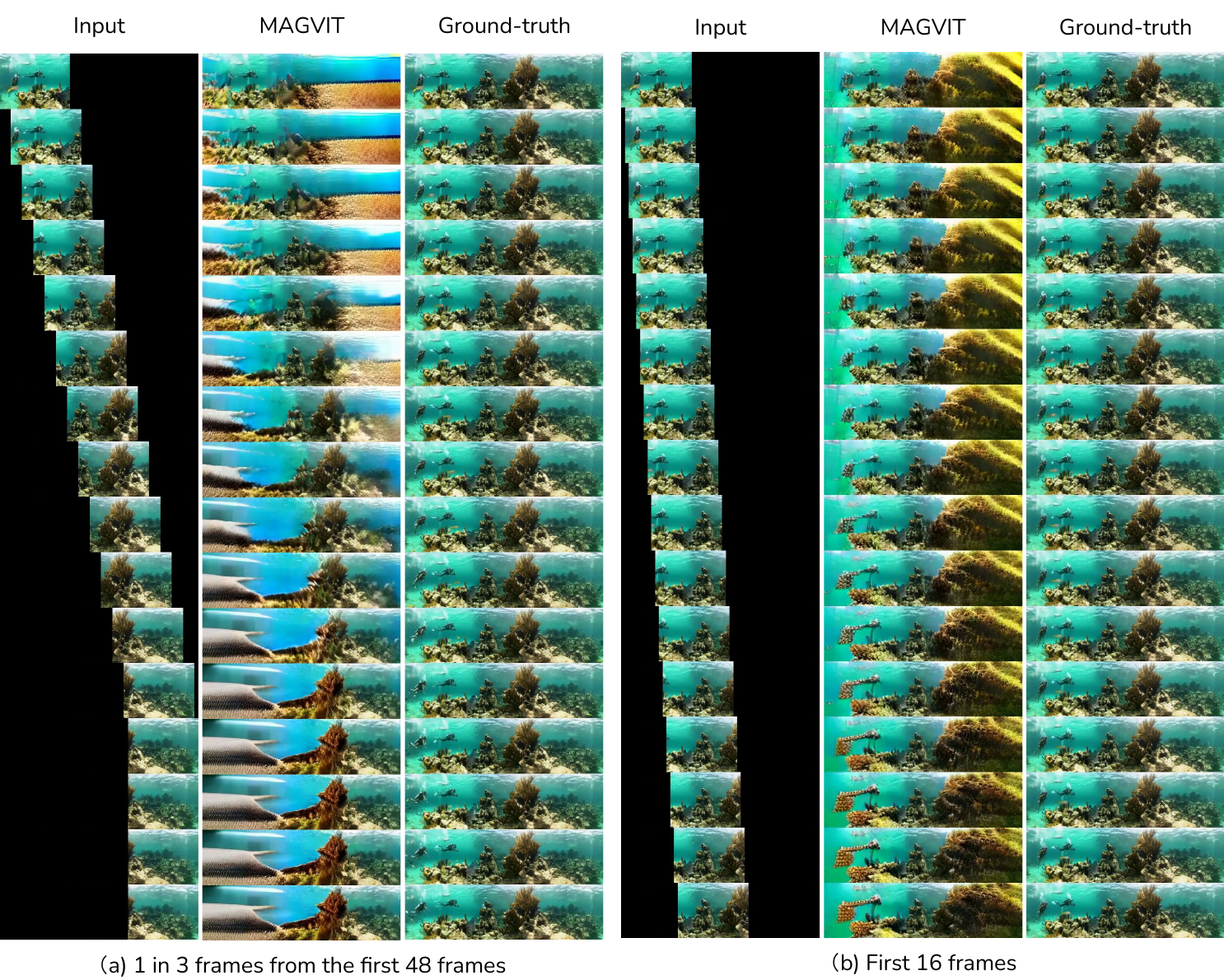}
    \vspace{-0.8em}
    \caption{MAGVIT baseline in two configurations. MAGVIT operates on a context window of 16 frames. Here we show with an example two different ways we select the 16-frame subset from the synthetic videos. One way is to take 1 in 3 frames from the first 48 frames (left) to span the full scene and minimize hallucination. We compare this version with linear interpolation and our results in Fig. 5 and Table 1. Since the fast panning motion from subsampling might be challenging for MAGVIT, we additionally show a version where we run on the first 16 frames of the videos, which is closer to the panorama outpainting setting in the original work (right). The camera pans slowly with large frame-to-frame overlap, at the expense of observing a portion of the scene. In both settings, MAGVIT struggles to synthesize consistent and realistic content at spatial locations far from the observed input window.}
    \vspace{-1em}
    \label{fig:supp_magvit}
\end{figure*}
\begin{figure*}[h]
    \centering
    \includegraphics[width=0.98\linewidth]{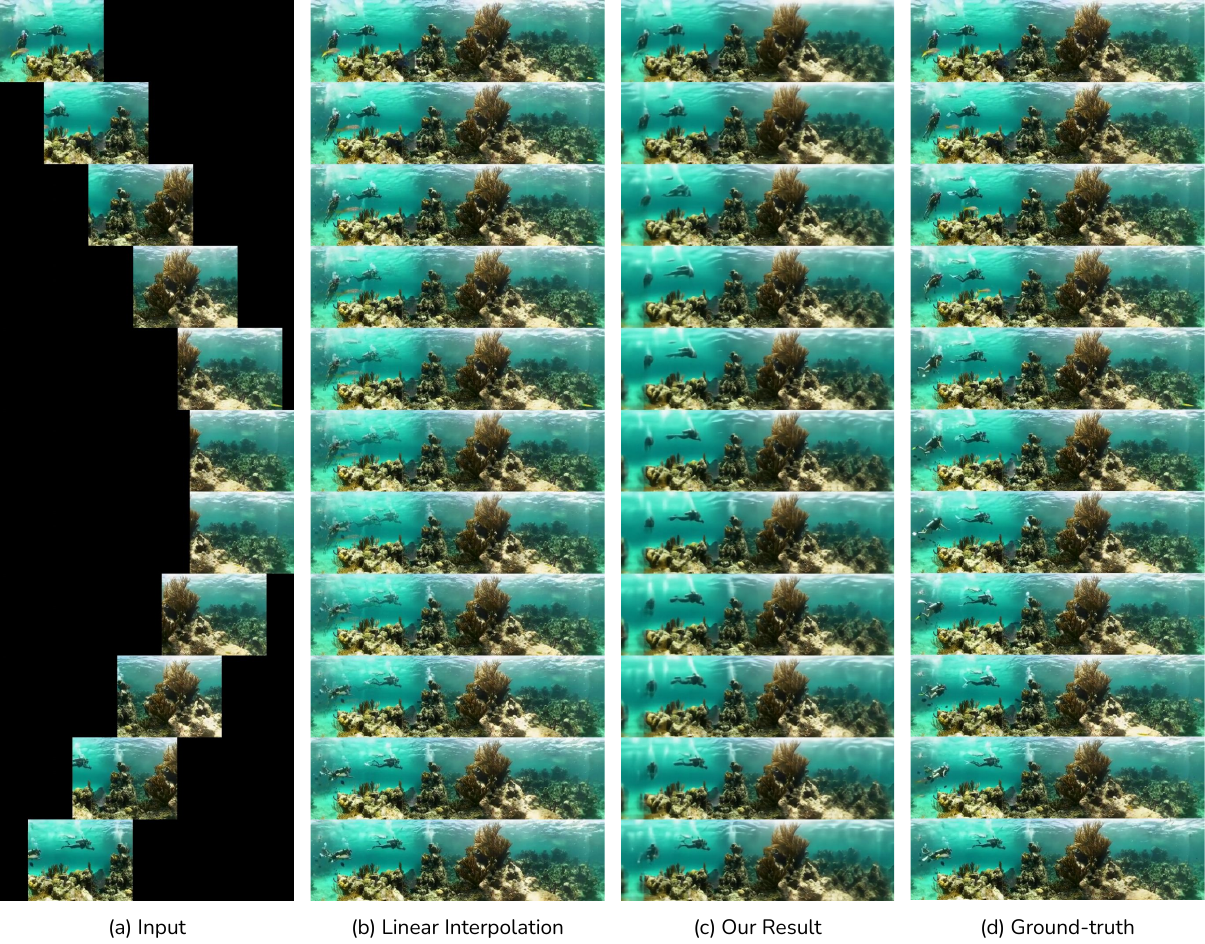}
    \vspace{-1em}
    \caption{Comparison with linear interpolation baseline (88-frame video, showing 1 in 8 frames, 11 frames total). Given an input video (a) with a left-right-left pan, the linear interpolation result (b) have degenerate motion, \eg divers on the left cross-fade in the synthesized regions, while our result (c) have smoothly interpolated motion and look consistent with the ground-truth (d).}
    \vspace{-1em}
    \label{fig:supp_comparisons}
\end{figure*}
\begin{figure*}[h]
    
    \centering
    \includegraphics[width=0.72\linewidth]{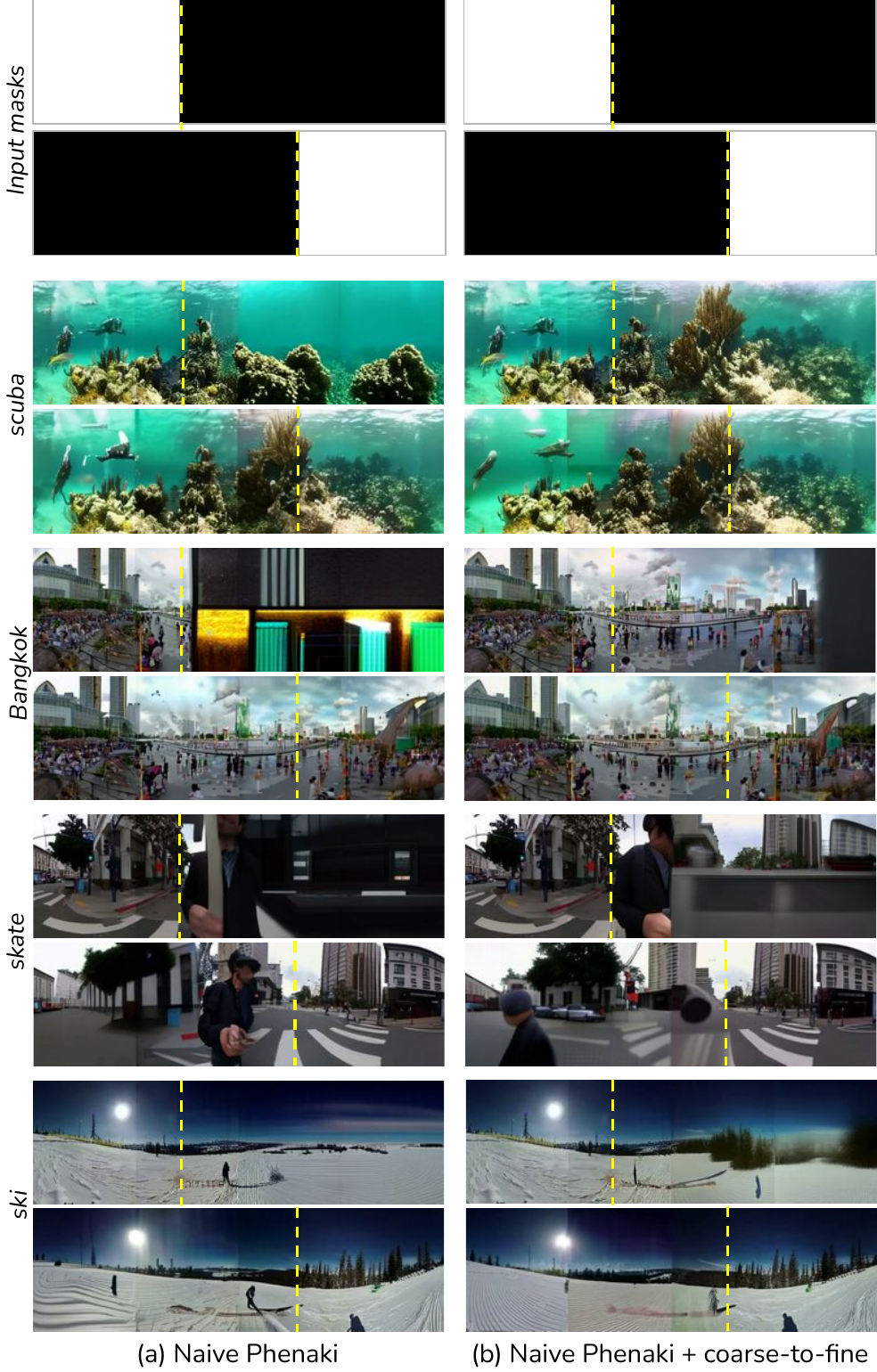}
    \caption{Ablation of the method. Left: a ``naive'' Phenaki result of completing the entire panorama without coarse-to-fine, spatial aggregation, or decoder finetuning. Right: our method without spatial aggregation or finetuning. Compare with our full method and ground-truth in Fig. 5. 
    Yellow dotted lines visualize the boundary of the visible input region.}
    \vspace{-1em}
    \label{fig:supp_ablation}
\end{figure*}

\section{Baseline/Evaluation details}
\subsection{MAGVIT}
For the MAGVIT baseline, we adopt the same panorama generation procedure as the original work: given the visible 64x128 center region, the model outpaints 32 pixels on both sides to obtain a 128x128 result. This procedure is repeated multiple times on both sides until the desired width is achieved. We resize the synthetic input panning video to a height of 128 pixels, preserving the aspect ratio, and outpaint on both sides. We then apply stabilization to the outpainted panning video by cropping accordingly.

Since the MAGVIT model operates on 16-frame videos, we subsample every 3rd frame from the first 48 frames of each synthetic video to obtain 16 input frames. This sampling allows the model to observe every spatial location of the input panorama, as the camera pans from the leftmost window to the rightmost window within the first 48 frames. We report numbers in Table 1 on the 16-frame subset. Since the large camera motion from subsampling may pose a challenge for the MAGVIT model, we additionally show MAGVIT outpainting results on the first 16 frames of each synthetic video in our supplementary webpage. This reduces camera motion at the expense of the model's only being able to observe a small portion of the scene. As seen in Fig. \ref{fig:supp_magvit}, MAGVIT still struggles to outpaint consistent and realistic content at spatial locations far from the observed input window.

\vspace{-0.4em}
\subsection{Linear interpolation}
In Fig. \ref{fig:supp_comparisons}, we show a comparison between linear interpolation and our results over a sequence of frames. Linear interpolation has obvious motion artifacts, for example, the divers on the left cross-fade between two observations, while in our result the divers are consistent with the ground-truth video and have plausible motion trajectories.

\subsection{Disentangled static/dynamic evaluation}
For pixel-level metrics (\ie PSNR, EPE), we split the inpainted regions into static and dynamic regions for disentangled evaluation. We determine static/dynamic regions by calculating optical flow and thresholding by flow magnitude of 0.2 pixels. For video clips with a moving camera, most pixels are categorized as dynamic.

\section{Phenaki Ablations}

We analyze two main ablations of our token-based method: ``naive'' Phenaki with no temporal coarse-to-fine synthesis, and Phenaki with temporal coarse-to-fine but no alignment or finetuning. Ablation results on the same videos as Fig. 9 
are shown in Fig.~\ref{fig:supp_ablation}.

\paragraph{Naive Phenaki.} The most straightforward way to complete a video volume using Phenaki is to apply the model in 11-frame, $160\times96$ sliding windows from the beginning to end of the video. 
As seen in Fig.~\ref{fig:supp_ablation}a, this baseline lacks temporal consistency as static inpainted regions differ from later observations of these regions (\eg the landscape in the right half of the panorama, for ``scuba'' and ``ski''). Furthermore, due to Phenaki's causal training (see Sec.~\ref{sec:baselevel}), severe artifacts are exhibited in regions that are unseen at the start of the video (denoted in orange in Fig.~\ref{fig:fwd_bwd_diagram}), particularly for ``Bangkok'' and ``skate''.

\paragraph{Phenaki with coarse-to-fine.} This ablation applies the coarse-to-fine synthesis and merging, but does not apply flow alignment or decoder finetuning. While an improvement over naive Phenaki, the results still contain large artifacts (\eg Fig.~\ref{fig:supp_ablation}b, ``Bangkok'' and ``skate''), as well as landscape inconsistencies (``ski''). Our full model avoids these artifacts and produces results consistent with the input video.

\end{document}